# Interpretable Machine Learning for Cognitive Aging: Handling Missing Data and Uncovering Social Determinant

Xi Mao, PhD[a1], Zhendong Wang, PhD [b1], Jingyu Li, MS[c], Lingchao Mao, PhD[c], Utibe Essien, MD[d], Hairong Wang, PhD[e]*, Xuelei Sherry Ni, PhD[f*]

a. Department of Economics, Robert C. Vackar College of Business & Entrepreneurship, University of Texas Rio Grande Valley, Edinburg, TX, USA
b. Shandong Mental Health Center, Jinan, Shandong Province, China.
c. H. Milton Stewart School of Industrial and Systems Engineering, Georgia Institute of Technology, Atlanta, GA, USA
d. Department of Medicine, University of California Los Angles Health, Los Angles, CA, USA
e. Operations Research and Industrial Engineering, Cockrell School of Engineering, University of Texas Austin, Austin, TX, USA
f. School of Data Science and Analytics, College of Computing and Software Engineering, Kennesaw State University, Kennesaw, GA, USA

[1] **co-first authors**

**\* to whom correspondence should be addressed:**

**e\***

**Telephone:** *404-820-2592*
**Email:** *hairong.wang@austin.utexas.edu*
**Address:**
*204 E. Dean Keeton Street*
*ETC II 5.120*
*Austin, Texas 78712*
*USA*

**f\***

**Telephone:** *470-578-2251*
**Email:** *sni@kennesaw.edu*
**Address:**
*680 Arntson Drive, Kennesaw State University*
*Room 3411C, MD 9044*
*Marietta, GA 30060*
*USA*

**Abstract**

Early detection of Alzheimer's disease (AD) is critical because its neurodegenerative effects are irreversible, and neuropathologic change as well as modifiable social–behavioral risk factors accumulate years before clinical diagnosis. Identifying higher-risk individuals earlier enables prevention, timely care, and more equitable resource allocation. We study prediction of cognitive performance from social determinants of health (SDOH) using the NIH NIA supported PREPARE Challenge Phase 2 dataset derived from the nationally representative Mex-Cog cohort from the 2003 and 2012 Mexican Health and Aging Study (MHAS).

**Data:** The target is a validated composite cognitive score in seven domains: orientation, immediate and delayed memory, attention, language, constructional praxis, and executive function, which derived from 2021 and 2016 MHAS. We curated features across demographic, socioeconomic, health status, lifestyle, psychosocial, and healthcare access.

**Methodology:** These feature domains were selected to capture the multidimensional nature of social and behavioral influences on cognitive aging. Substantial missingness was addressed with a singular value decomposition (SVD)-based data imputation pipeline that treats continuous and categorical variables separately. This approach leverages latent correlations among features to recover missing values and offers a strong balance between statistical reliability and scalability. Through comparative evaluation of multiple methods, XGBoost was selected as the default prediction model due to its superior performance.

**Results:** Our results show that the proposed framework outperforms existing methods and achieves better predictive accuracy than the top results reported on the data challenge leaderboard. To better understand the relationship between input features and the composite cognitive score, we conducted a thorough post hoc analysis of the top contributing features, examining the mechanism by which these features are associated with cognitive scores. The study further stratified the analysis by age group based on SHAP analysis to explore whether the most predictive features differ across life stages. The proposed pipeline outperformed the state-of-the-art methods, demonstrating robustness, interpretability, and computational efficiency, and underscoring its potential as a practical modeling strategy for datasets with substantial missingness in both continuous and categorical features.

**Discussion:** Our findings underscore the significance of flooring material as more than a mere housing characteristic – it serves as a powerful proxy for deeper structural determinants of health. Its predictive strength lies in its ability to reflect socioeconomic status (SES), environmental exposures, and access to healthcare, all of which shape long-term health outcomes. This reinforces the idea that seemingly minor household features can encapsulate broader systemic inequities. Alongside flooring, other key predictors such as SES, age, lifestyle interventions (e.g., cognitive and leisure activities), social interaction, sleep quality, stress, and BMI collectively highlight the multifactorial nature of SDOH trajectories.

## 1. Introduction

Alzheimer's Disease and Alzheimer's disease and related dementias (AD/ADRD) impose substantial health and economic burdens through costly late-stage care, rising prevalence, and intensive caregiving demands. Earlier detection can meaningfully improve health and healthcare decisions and outcomes by enabling timely intervention and slowing progression.

This study investigated cognitive performance by leveraging social determinants of health (SDOH) using data from the NIH NIA-supported PREPARE Challenge Phase 2, derived from the nationally representative Mex-Cog cohort within the 2003 and 2012 Mexican Health and Aging Study. The analysis included a broad range of features across demographic, socioeconomic, health status, lifestyle, psychosocial domains, and healthcare access.

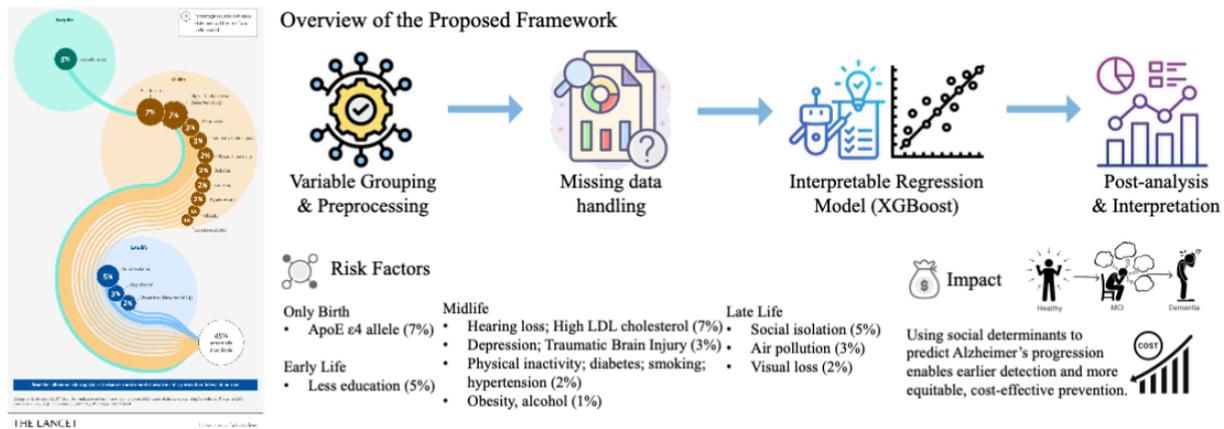

**Figure 1: Overview of the Proposed Work**

The conceptual foundation of this study draws on evidence from *The Lancet* Commission, which highlights the cumulative influence of modifiable risk factors across the life course on dementia development. The visual representation underscores four critical stages, birth, early life, midlife, and late life, where exposures such as limited education, hearing loss, high cholesterol, and social isolation contribute to Alzheimer's disease risk. These factors, quantified by their relative impact, emphasize the need for preventive strategies that span decades rather than focusing solely on late-life interventions (Figure 1).

Building on this evidence, the proposed analytical framework follows a structured, four-step process (Figure 1). First, variables are grouped and preprocessed to ensure coherent integration of social determinants and clinical data. Second, missing data are systematically addressed to maintain analytical robustness. Third, an interpretable regression model based on XGBoost is employed, balancing predictive accuracy with transparency in feature contributions. Finally, post-analysis and interpretation synthesize model outputs into actionable insights, enabling researchers and clinicians to understand the drivers of disease progression.

Consistent with existing literature, the study found that factors such as socioeconomic status (e.g., education, mother's education, occupation, employment status, hours worked, and household income), sleep (e.g., restless sleep and sleep duration), body mass index, communication (e.g., frequency of social activities, visits with friends and relatives, and number of living children), participation in cognitively stimulating entertainments, and perceived stress are all associated with MCI and AD.

A particularly novel finding from this research was the significance of flooring material as a predictive feature for MCI and ADRD detection. Rather than being a mere aspect of housing, flooring material proved to be a compact surrogate marker for various latent constructs, such as poverty, access to care, the physical living environment, and psychosocial well-being. Flooring material encapsulates housing quality or wealth, which includes broader structural inequalities and environmental exposures that have tangible effects on downstream health trajectories. This makes it one of the most informative and impactful features in the prediction model for cognitive impairment within the population in Mex-Cog.

## 2. Data Acquisition and Study Population

We used the dataset provided by the NIH Alzheimer's Disease Social Determinants of Health (SDOH) Challenge, hosted by DrivenData[1]. The dataset was derived from the Mexican Health and Aging Study (MHAS) Cognitive Aging Ancillary Study (Mex-Cog), a nationally representative, longitudinal survey of older adults in Mexico.

This publicly available dataset was constructed to facilitate research on the impact of social and behavioral factors on cognitive aging and Alzheimer's disease progression. It includes over 3000 de-identified participant records and includes a broad set of survey-based variables in 2003 and 2012, capturing demographic characteristics (e.g., age, sex, race/ethnicity, education), health-related behaviors (e.g., smoking, alcohol use, physical activity), clinical indicators (e.g., BMI and chronic conditions), and cognitive or functional assessments. The prediction target is a composite cognitive score computed from in-depth cognitive assessments administered in person as part of Mex-Cog. This score aggregates performance across seven cognitive domains: orientation, immediate memory, delayed memory, attention, language, constructional praxis, and executive function, with a theoretical maximum of 384 points. Higher scores indicate better cognitive function. Scores are available for two outcome years (2016 and 2021), enabling prediction at two time points using features collected in 2003 and 2012. The dataset and its accompanying documentation are available for public download from the DrivenData competition portal.

The dataset exhibited a high rate of missingness across variables, a common challenge in large observational studies integrating socioeconomic and health data. To address this, we applied an imputation strategy based on singular value decomposition (SVD), leveraging matrix completion techniques to estimate missing entries under the assumption of low-rank structure in the data. This approach exploits latent correlations among features to recover incomplete values and provides a balance between statistical robustness and computational efficiency. The detailed imputation process is described in the next subsection.

## 3. Method

### 3.1 Handling of Missing Data via SVD-based Matrix Completion

To address missing values in the dataset, we employed a low-rank matrix completion approach centered around a data-driven method for selecting the optimal hard threshold in singular value decomposition. This method was applied separately to continuous and categorical variables to account for their distinct characteristics. The process began with data preprocessing: categorical variables were converted into one-hot encoded form with missing entries set to zero (Yu et al., 2022), while continuous variables were normalized using z-score normalization before their missing entries were likewise initialized to zero.

The core of the method involved an iterative imputation process based on SVD. In each iteration, the incomplete data matrix was first initialized with current estimates of the missing entries and then decomposed via singular value decomposition (SVD). A hard-thresholding operation was applied to the resulting singular values: all singular values greater than a specified threshold $\lambda$ were retained, while those less than or equal to $\lambda$ were directly set to zero. The matrix was subsequently reconstructed using the thresholded singular values, and the missing entries were updated using the corresponding values from this reconstructed matrix. All observed entries were then reset to their known values before the next iteration. This iterative process continued until the root mean squared error (RMSE) between the reconstructed and observed values fell below a predefined tolerance threshold, ensuring convergence to a stable low-rank approximation (Candes & Recht, 2012) (Candes & Plan, 2010).

---

[1]https://www.drivendata.org/competitions/300/competition-nih-alzheimers-sdoh-2    4

Determining the optimal hard threshold λ was a critical aspect of the method. We adopted a cross-validation-based search strategy: first, a grid of candidate thresholds covering the range of the initial singular values was generated. Then, for each candidate threshold $\lambda$, K-fold cross-validation was performed. This involved randomly partitioning the known data into training and validation sets, performing the hard-thresholded SVD imputation using the training set and the current threshold λ, and calculating the mean squared error on the validation set. The candidate threshold that yielded the smallest average validation error was selected as the global optimal threshold $\lambda^*$. This approach provided an objective basis for threshold selection, consistent with the principles of robust model selection(Athey et al., 2021) (Candes & Plan, 2010).

After obtaining the optimal threshold and completing the iterative imputation for the entire dataset, post-processing was applied: the imputed results for categorical variables were converted back to the original categorical format by selecting the category corresponding to the dimension with the maximum value for each sample, while the imputed results for continuous variables were denormalized using the stored means and standard deviations to return to the original data scale. This adaptive low-rank approximation, achieved through hard thresholding, offers a rigorous and efficient solution for handling missing data.

**3.2 Machine Learning Model Specification**

We adopted XGBoost, a scalable and efficient gradient-boosted decision tree algorithm, as the core modeling framework to predict *composite_score* across different age groups. The selection of XGBoost was driven by the intrinsic challenges of this prediction task and the unique characteristics of the dataset. Specifically, the data comprise a mixture of continuous and categorical variables spanning demographic, behavioral, clinical, and socioeconomic domains. The dataset is also high-dimensional, with more than 100 features, some of which are correlated or sparsely populated. Moreover, the dataset contains noise introduced during missing data imputation, as the imputation process, while carefully designed, cannot fully eliminate uncertainty or reconstruct the original data distribution. These challenges render simple linear or shallow models insufficient, while making highly parametric models, such as deep neural networks, prone to overfitting given the moderate sample size.

Gradient-boosted decision trees have been shown to perform exceptionally well on tabular data, especially when variables are heterogeneous and nonlinearly related to the outcomes. XGBoost, in particular, is designed to optimize both speed and predictive performance by introducing several algorithmic innovations: (1) *regularization* via both $L_1$ and $L_2$ penalties, which enhances generalizability and controls model complexity; (2) *sparsity-awareness* allows the algorithm to handle missing or imputed values by automatically learning the optimal splitting directions for empty entries; (3) *tree boosting with shrinkage and column subsampling*, which improves robustness to noisy features and prevents overfitting; and (4) *efficient parallelization and* *scalability* for large or high-dimensional datasets (Chen & Guestrin, 2016) (Friedman, 2001) (Ke et al., 2017) (Lundberg et al., 2020). These properties make XGBoost particularly well-suited to our data structure, which contains a mix of numerical and categorical predictors with potential nonlinear interactions. Unlike neural networks, which require extensive feature normalization and large-scale training data, XGBoost can directly accommodate the raw feature structure while still capturing complex relationships.

In recent years, XGBoost has emerged as one of the most successful machine learning algorithms for tabular biomedical and epidemiological data. For instance, studies have shown that tree-based ensemble models consistently outperform deep learning models in clinical risk prediction tasks when dealing with heterogeneous features (Lundberg et al., 2018) (Shickel et al., 2017). Moreover, XGBoost has proven robust against moderate levels of noise and is widely regarded for its interpretability when combined with SHAP (SHapley Additive exPlanations) values (Lundberg & Lee, 2017) (Lundberg et al., 2020) (Athanasiou et al., 2020) (Lu et al., 2022). Given these advantages, XGBoost has become a widely adopted algorithm in the medical and healthcare domain, where it has been extensively applied to clinical risk prediction, disease diagnosis, and patient outcome modeling. Its strong performance on structured, heterogeneous datasets has established it as a preferred choice for both research and practical applications in these fields.

## 4. Results

Using features collected in 2003 and 2012, we trained a gradient-boosted decision tree regressor (XGBoost) to predict the composite score. To better capture the temporal dynamics between feature changes and the target outcome, we included an additional feature representing the time interval between feature collection and the target score. Instead of directly using both ages measured in 2003 and 2012, we excluded the age recorded in 2003 (age_03), as the target already reflected the passage of time. The age measured in 2012 was retained to serve as a patient-specific age feature and to improve the predictive power of the model.
Model performance was assessed via 10-fold cross-validation. The model achieved an average RMSE of 36.65 with a standard deviation of 1.47 across folds, indicating stable out-of-sample performance.

To identify the best-performing model, we explore a diverse set of commonly used regression models, including XGBoost, LightGBM, CatBoost, Random Forest, Decision Tree, Linear Regression, and Support Vector Regression (SVR). These models were selected to represent a spectrum of learning paradigms, including gradient boosting methods, tree-based ensembles, linear models, and kernel-based methods. They are widely adopted in both academic and industrial settings due to their strong empirical performance on tabular data. In particular, tree-based models like XGBoost (Chen & Guestrin, 2016) and LightGBM (Ke et al., 2017) are known for their ability to capture complex feature interactions, while CatBoost (Prokhorenkova et al., 2018) further supports native categorical feature handling and missing value robustness. Random Forest and Decision Tree offer interpretable non-linear baselines, and SVR and Linear Regression provide contrastive perspectives from parametric and kernel-based learning.

We first evaluate model performance using our proposed framework, which employs an SVD-based imputation method to handle missing values prior to training. The imputed data is then used to train each of the candidate models under a consistent 10-fold cross-validation protocol. As shown in Figure 1, models based on gradient boosting, particularly XGBoost, outperform other approaches, achieving the lowest average RMSE with smallest standard deviation (std).

To assess the impact of the imputation strategy, we compared model performance under two distinct preprocessing pipelines. In the baseline setting (Figure 2b), missing values in numerical features were filled using median imputation, and categorical features were imputed with the most frequent category followed by one-hot encoding. In contrast, our proposed framework (Figure 2a) uses the SVD-based method to jointly reconstruct missing entries by exploiting low-rank correlations across features before model training. Results show that SVD-based imputation consistently improves or maintains performance across all models. The most significant gains are observed in Random Forest and in Decision Tree. On the other hand, for models like XGBoost, LightGBM, and CatBoost, the difference between the two imputation strategies is relatively small, suggesting that these gradient-boosted models are more robust to moderate levels of imputation noise or error. These findings demonstrate that while advanced boosting models can tolerate simpler imputation strategies, models without internal missing data handling mechanisms, such as Random Forest and Linear Regression, can significantly benefit from more principled, structure-aware imputation methods.

In summary, models that natively handle missing values show modest yet consistent improvements when trained on SVD-imputed data, indicating that structure-aware imputation can still refine learning even for robust models. In contrast, models that do not support missing values demonstrate notable performance gains under the SVD-based imputation strategy, highlighting its effectiveness in improving data quality prior to training. These results highlight the effectiveness of integrating low-rank structure recovery with non-linear predictive models when facing incomplete high-dimensional data.

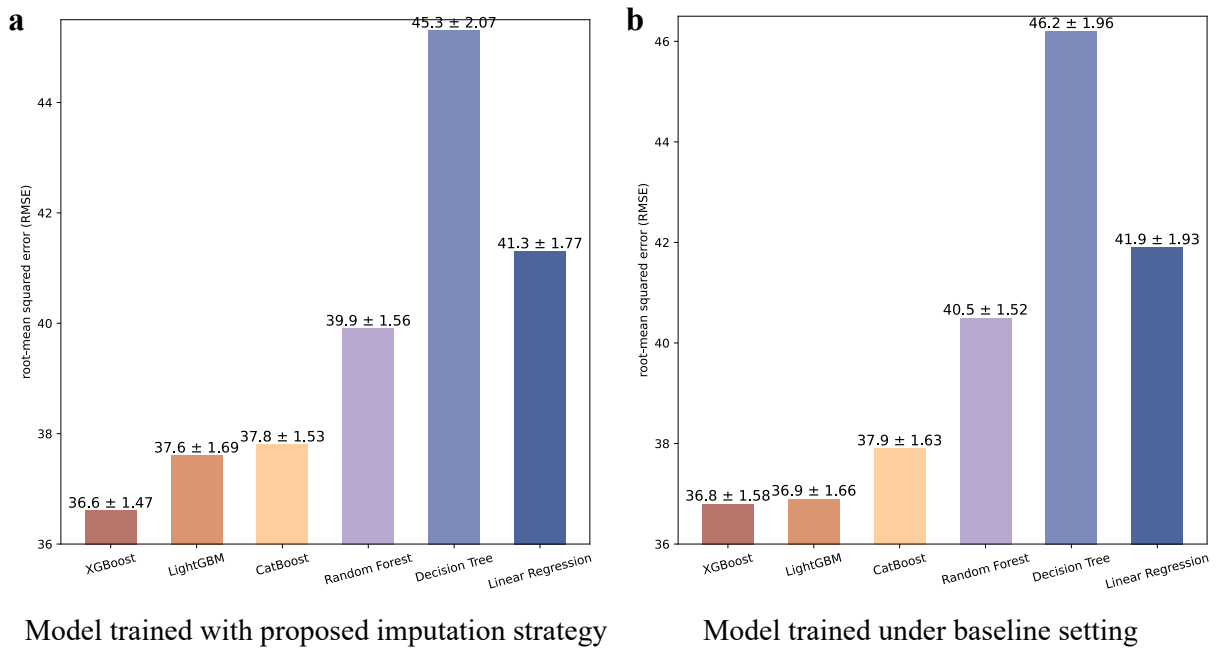

**Figure 2.** Model performance comparison under different missing data imputation strategies. Bar plots show the average root-mean squared error (RMSE) and standard deviation (±) across 10-fold cross-validation for six regression models. **(a)** Results using the proposed framework, where missing values are imputed using a singular value decomposition (SVD)-based approach prior to model training. **(b)** Results using a baseline strategy, where missing numerical values are imputed with the median and categorical variables with the mode followed by one-hot encoding.

## 4.1 Post Hoc Interpretation of Feature Importance Using SHAP

To better understand the model behavior and gain interpretability into the prediction process, we conducted a post hoc analysis by examining the top 20 most important features identified by the best-performing model (XGBoost), as shown in Figure 3 and the variable names are listed in Table 1. Feature importance was computed based on the SHAP values, allowing us to rank features by their relative contribution to the overall predictive performance. SHAP is a model-agnostic interpretability framework grounded in cooperative game theory, which attributes a unique importance value to each feature based on its marginal contribution to the prediction for each instance.

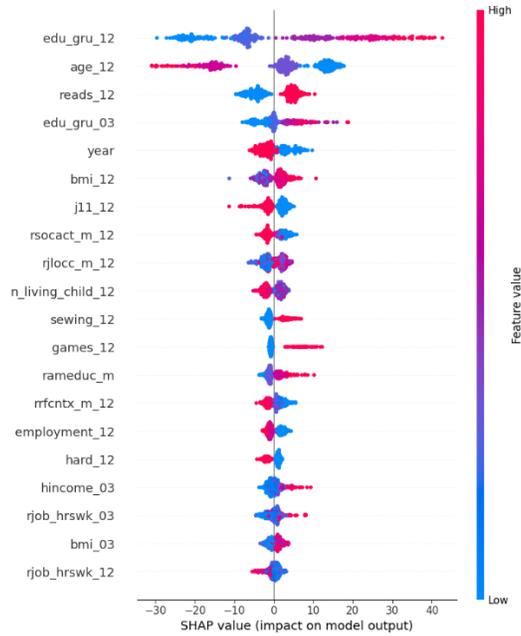

**Figure 3.** SHAP summary plot of feature importance for the full patient cohort. This plot shows the top 20 most important features driving model predictions, ranked by their mean absolute SHAP value. Each point represents an individual patient, with color indicating the feature value (red = high, blue = low), and horizontal position indicating the SHAP value (impact on model output). Features such as education level (edu_gru_12, edu_gru_03), age group (age_12), reading behavior (reads_12), and flooring material (j11_12) had the strongest influence on predicted outcomes. Positive SHAP values indicate contributions toward higher predicted composite scores, while negative values reflect contributions toward lower scores.

**Table 1.** Top 20 most important predictive features identified by SHAP values for the full cohort. The table lists the top 20 features ranked by their mean absolute SHAP values, indicating their contribution to the model's predictions. Each feature is accompanied by a brief description and its potential relevance to the outcome.

| Rank | Feature Name | Description | Potential Relevance |
|---|---|---|---|
| 1 | edu_gru_12 | Education level in 2012 (binned) | Cognitive reserve, SES indicator |
| 2 | age_12 | Age group in 2012 (binned) | Age is directly linked to health/cognition |
| 3 | restless_12 | Felt restless sleep most of the past week | Proxy for sleep quality and mental health |
| 4 | edu_gru_03 | Education level in 2003 (binned) | Earlier SES/cognitive reserve indicator |

| 5 | year | Survey year (2003 or 2012) | Captures temporal shift or time-dependent effect |
| 6 | bmi_12 | Body Mass Index category in 2012 | Physical health and metabolic risk |
| 7 | j11_12 | Type of flooring material at home | Proxy for housing quality / economic status |
| 8 | rsocact_m_12 | Frequency of social activities in 2012 | Reflects social engagement and interaction |
| 9 | rjlocc_m_12 | Longest held job category | Occupation type affects exposure, cognition |
| 10 | n_living_child_12 | Number of living children (binned) | Family support or caregiving burden |
| 11 | sewing_12 | Time spent sewing or doing crafts | Fine motor skills and leisure cognitive activity |
| 12 | games_12 | Time spent playing puzzles or games | Cognitive engagement |
| 13 | rameduc_m | Mother's education level | Familial SES, early life influence |
| 14 | rfcntrx_m_12 | Frequency of seeing friends/relatives in 2012 | Social support network |
| 15 | employment_12 | Current employment status | Functional status and economic activity |
| 16 | hard_12 | Felt everything was an effort (past week) | Proxy for fatigue, depression |
| 17 | hincome_03 | Household income in 2003 | Long-term socioeconomic status |
| 18 | rsocact_m_03 | Frequency of social activities in 2003 | Earlier life social involvement |
| 19 | bmi_03 | Body Mass Index category in 2003 | Baseline physical health |
| 20 | rjob_hrswk_12 | Weekly hours worked at main job in 2012 | Labor intensity and physical/cognitive strain |

## 4.2 Performance Stratified by Age Group

To better understand the model's generalizability across different patient populations, we conducted a stratified performance analysis by age group. Age is a critical factor in health trajectories, functional status, and data completeness, and may affect both the distribution of input features and the predictability of the composite outcome. Therefore, we divided the cohort into five age groups: Below 49, 50–59, 60–69, 70–79, and Above 80.

As shown in Figure 4, prediction performance varies considerably across age groups. The lowest RMSE was observed in the 50–59 group (35.1 ± 2.03), followed closely by 60–69 (36.5 ± 3.51), indicating high model accuracy in mid-life adults. In contrast, individuals younger than 49 exhibited the highest prediction error (44.5 ± 11.03), with notably large variability. The oldest group (80+) also showed elevated RMSE (41.6 ± 3.90), though with narrower uncertainty.

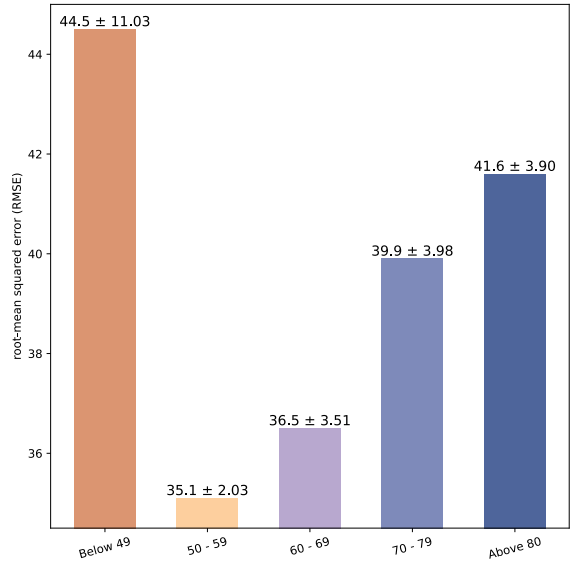

**Figure 4.** Model performance across age groups. The bar plot shows the root mean squared error (RMSE) of model predictions stratified by age group.

These discrepancies may be driven by several factors. For younger individuals, greater heterogeneity in health status and lifestyle factors, combined with less accumulated clinical history, may lead to noisier data and weaker signal. For older adults (especially 80+), higher error may reflect complex comorbidities, measurement errors, or missingness patterns not fully captured by the model. In contrast, middle-aged groups likely benefit from more stable health trajectories, richer observed data, and clearer feature-outcome relationships, leading to better model performance.

To further explore the age-related differences in model behavior, we analyzed the top contributing features in Table 2 for each age group using SHAP summary plots (Figure 5). This comparison reveals both shared and distinct patterns of feature importance across the life course.

In the Below 49 group, model predictions were largely influenced by variables reflecting early-life socioeconomic status (SES) and baseline health history, including education level in 2003 (edu_gru_03), Household income in 2003 (hincome_03), and mother's education (rameduc_m). This aligns with the notion that in younger individuals, long-term outcomes may still be shaped by foundational background and early exposures, as their current clinical profiles are less developed or less predictive.

In contrast, individuals aged 50–59 and 60–69 exhibited top features that reflect current household structure, functional health, and behavioral engagement, such as time spent on readings on 2012 (reads_12), flooring material in 2012 (j11_12), social activity in 2012 (rsocact_m_12), Number of living children in 2012 (n_living_child_12), and social contact frequency (rfcntx_m_12). These mid-life groups showed high prediction accuracy (low RMSE), likely because the model could leverage both social and functional predictors that are highly informative in this transitional phase between middle and older age.

Among the older adults (70–79 and 80+), the most important features shifted toward functional limitations, health utilization, and retirement-related factors, such as spouse earnings in 2012 (searnings_12), BMI in 2012 (bmi_12), self-rated global health (glob_hlth_12), and weekly hours worked at main job in 2012 (rjob_hrswk_12). Notably, Education level (edu_gru_12 and edu_gru_03) remained important across all groups, but the relative contribution of health behavior and income-related features increased with age. For instance, in the oldest group (Above 80), predictors like time spent on watching TV (tv_12), time to talk on the phone or send message (comms_tel_comp_12), and hospitalization gained prominence, possibly reflecting reduced mobility and increased healthcare dependence.

These differences suggest that the model adapts to different predictors depending on the life stage, with early-life factors dominating in youth, social and behavioral variables driving predictions in middle age, and health status and care access becoming more influential in older adulthood. This also highlights the need for age-aware feature interpretation in longitudinal predictive modeling.

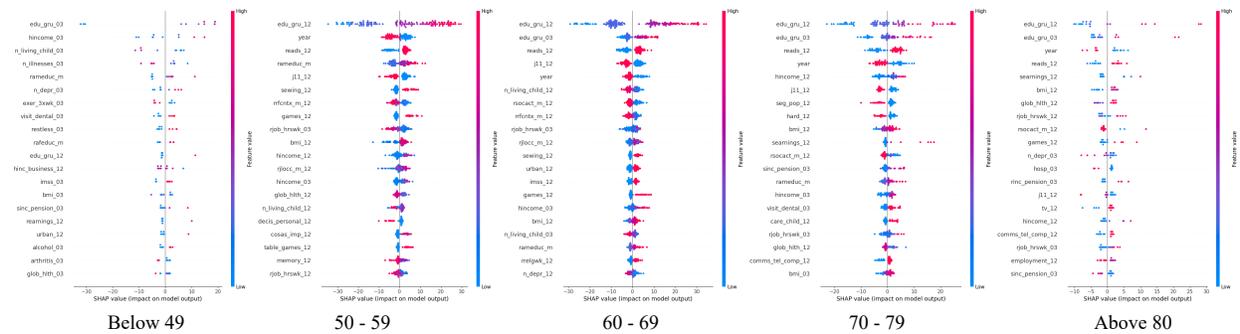

| Below 49 | 50 - 59 | 60 - 69 | 70 - 79 | Above 80 |

**Figure 5.** Top predictive features across different age groups based on SHAP analysis. SHAP summary plots show the top 20 most influential features contributing to model predictions for each age group: Below 49, 50–59, 60–69, 70–79, and Above 80.

**Table 2.** Top 5 most predictive features by age group based on SHAP analysis. This table summarizes the five most influential features for each age group, as identified by SHAP values from the XGBoost model. The features reflect a diverse range of domains, including socioeconomic status (e.g., education, income), health indicators (e.g., BMI, self-reported illnesses), lifestyle and cognitive engagement (e.g., reading habits, social activities), and contextual variables (e.g., type of flooring as a proxy for housing quality). Across all age groups, education level in 2012 (edu_gru_12) emerges as a dominant predictor. Younger groups (e.g., Below 49) show greater influence from early-life and familial SES markers (e.g., mother's education, childhood household income), while older groups (e.g., Above 80) are more affected by current health and lifestyle factors. This age-stratified analysis highlights how different feature types contribute variably to cognitive health across the lifespan.

| Age Group | Top 5 SHAP Features | Description |
| --- | --- | --- |
| Below 49 | edu_gru_03 | Education level in 2003 (early-life SES proxy) |
| | hincome_03 | Household income in 2003 |
| | n_living_child_03 | Number of living children (2003) |
| | n_illnesses_03 | Number of self-reported illnesses |
| | rameduc_m | Mother's education level |
| 50–59 | edu_gru_12 | Education level in 2012 |
| | reads_12 | Reads books, magazines, newspapers |

| Age Group | Top 5 SHAP Features | Description |
|---|---|---|
| | j11_12 | Type of flooring material (housing quality proxy) |
| | n_living_child_12 | Number of living children (2012) |
| | rsocact_m_12 | Frequency of social activities |
| **60–69** | edu_gru_12 | Education level in 2012 |
| | reads_12 | Reads books, magazines, newspapers |
| | year | Survey year |
| | hincome_12 | Household income in 2012 |
| | j11_12 | Type of flooring material |
| **70–79** | edu_gru_12 | Education level in 2012 |
| | hincome_12 | Household income in 2012 |
| | bmi_12 | Body mass index category |
| | searnings_12 | Spouse's earnings |
| | j11_12 | Type of flooring material |
| **Above 80** | edu_gru_12 | Education level in 2012 |
| | reads_12 | Reads books, magazines, newspapers |
| | searnings_12 | Spouse's earnings |
| | bmi_12 | Body mass index category |
| | glob_hlth_12 | Self-rated global health |

## 5. Discussion

### 5.1 Significance of the Analytical Framework

The analytical framework developed in this study offers several key strengths. First, by combining low-rank SVD-based imputation with gradient-boost ensemble learning, the method achieves both robustness to data missingness and high predictive performance – a rare balance in SDOH-based epidemiologic modeling. This hybrid design enables extraction of latent social and behavioral patterns often obscured by incomplete data. Second, the use of cross-validated hard-threshold selection ensures the imputation process remain data-driven and avoids overfitting (Candes and Plan, 2010; Athey et al. 2021). Third, model interpretability is achieved through SHAP-based post hoc analysis, which quantifies feature-level contributions and supports transparent reasoning for clinical or policy translation (Lundberg and Lee, 2017). Finally, the framework's scalability and modular design make it readily adaptable for other population-based datasets with similar missingness or mixed data structures, supporting reproducible, multi-cohort learning paradigms.

### 5.2 Why type of flooring material important?

Among the top predictive features identified by SHAP analysis is type of flooring material (j11_12), which captures the housing quality in the respondent's residence in 2012. While this may seem like a superficial household attribute, previous studies evidence that type of flooring material as a highly informative proxy for housing quality, SES, wealth, and broader health-related vulnerabilities (Legge et al.,2023) (Vyas and Kumaranayake, 2026).

In Figure 6, SHAP values show that flooring material contributes substantially to prediction accuracy. To understand why, we examined its bivariate associations with other features. The Pearson correlation analysis reveals that flooring material is moderately associated with variables linked to education level (edu_gru_03, edu_gru_12), urbanicity (urban_03, urban_12), and mother's education (rameduc_m), all with negative correlations (r ≈ −0.2), indicating that poorer flooring material or housing quality associate with lower education and rural residence. Conversely, flooring material shows positive correlations with features such as global health status (glob_hlth_12), social contact frequency (rfcntx_m_12), and public insurance enrollment (seg_pop_12), suggesting that flooring type also reflects aspects of environmental, social and healthcare access.

These patterns are reinforced by the Cramér's V results, which show strong associations (V ≥ 0.5) between flooring material and several key SES variables, including urban residence (0.76), insurance coverage (0.75), household income (hincome_12, hinc_cap_12), and employment status (0.66). In addition, Chi-square tests confirm that flooring material is highly significantly associated (p < 1e-25) with a broad range of variables including life satisfaction, health status, depressive symptoms, and household economic indicators.

The combined findings indicate that flooring material is a concise indicator for underlying factors like poverty, healthcare access, living conditions, and SES. Rather than being a trivial housing detail, flooring material encapsulates broader structural inequities and environmental exposures that have measurable impacts on downstream health trajectories, making it one of the most informative features in our prediction model.

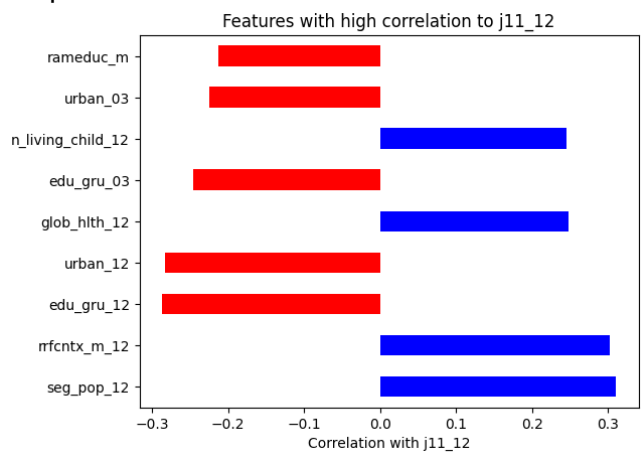

**Figure 6.** Top features with highest Pearson correlation to flooring material (j11_12). This bar plot shows the features most strongly correlated with flooring material, which serves as a proxy for household flooring quality and, by extension, socioeconomic status. Red bars indicate negative correlations, while blue bars indicate positive correlations. Notably, higher levels of education (edu_gru_12, edu_gru_03), global health status (glob_hlth_12), and social connectedness (rfcntx_m_12, seg_pop_12) are positively associated with better flooring conditions, whereas rural residence (urban_03, urban_12) and lower mother's education (rameduc_m) show negative associations. This highlights the role of flooring material as an environmental and SES-sensitive feature.

### 5.3 Other important features

#### 5.3.1 Socioeconomics Status (SES)

Other than the flooring material, SHAP values also show that higher educational attainment demonstrates a dual role in Alzheimer's disease and related dementias (ADRD). While educational attainment broadly correlates with reduced ADRD risk through cognitive reserve mechanisms. Better-educated individuals often report a higher cognitive reserve relative to others (Stern,2012). They also have subjective cognitive concerns earlier due to heightened self-awareness and greater tolerance to amyloid burden in clinical stages of mild cognitive impairment (MCI) and AD (Gonneaud,2020). Thus, higher-educated people have more cognitive resilience ability.

Besides of education, there is a rich literature introduce the correlation between other Socioeconomic Status factors, namely, occupation and income, is consistently linked to risk of MCI and AD. Individuals with limited household educational attainment, low household income, or unstable employment experience higher rates of cognitive decline, with disadvantaged groups especially affected (Livingston et al., 2020) (Zhang et al., 2022).

### 5.3.2 Age

According to our results, age is also a significant non-modifiable risk factor for Alzheimer's disease and related dementias (ADRD), with prevalence doubling every five years after age 65. Nearly half of adults aged 85+ develop dementia (Solane et al., 2002). Age-related mechanisms in ADRD include chronic inflammation, vascular dysfunction, and amyloid/tau accumulation. The APOE ε4 gene is a well-established biomarker for increased risk of Alzheimer's and mild cognitive impairment. APOE ε4's presence amplifies the effect of modifiable risk factors such as diabetes (Rashtchian et al., 2024).

Age can also affect ADRD associated with other risk factors in different stages of life. Such as, midlife (40–64 years) cardiovascular and metabolic factors significantly elevate late-life dementia risk. Hypertension, smoking, and diabetes in midlife increase dementia risk by 20–46%, with a dose-dependent relationship observed for cumulative risk factors (Whitmer et al., 2005). Conversely, late-life (≥65 years) risks shift toward social isolation, physical inactivity, and chronic conditions like depression, which accelerate cognitive decline. Lyu et al. (2024) show that social isolation in older adults is linked to a 41% higher ADRD risk.

### 5.3.3 Lifestyle Interventions (Cognitive and Leisure Activities)

Though higher education reduces dementia risk as aging, it loses protective effects in the oldest-old (≥85 years). Similarly, lifestyle interventions like cognitive activities (e.g., puzzles, reading, crafts) and social engagement show greater benefits in younger seniors (65–80 years) compared to older cohorts. Structured cognitive activities like puzzles and crosswords are theorized to enhance cognitive reserve. The mix effect of social crafting groups improves participation and self-efficacy in MCI patients, but direct evidence linking sewing, embroidery, or knitting to ADRD risk reduction is currently limited. Participation in cognitive and leisure activities, such as puzzles, crafts, and games, slows decline, improves memory, and reduces MCI risk (Akbaraly et al., 2009) (Wilson et al., 2002) (Krell-Roesch et al., 2019) (Verghese et al., 2006). Individuals engage in cognitively leisure activities that boost cognitive reserve and memory, providing resilience to dementia pathology with a lower risk of development of amnestic mild cognitive impairment, even after excluding individuals at early stages of dementia.

### 5.3.4 Communication and Social Interaction

Late-life social isolation, infrequent interpersonal communication, and low engagement in social activities substantially increase the risk of cognitive decline and dementia. Social connectivity, including frequency of friend visits and family interaction, offers notable protection against MCI and ADRD (Kuiper et al., 2015) (Livingston et al., 2024) (Myers et al., 2025) (Sutin et al., 2023). Thus, frequent social interactions significantly slow cognitive decline. Weekly engagement with family or friends correlates with slower memory decline, while consistent participation in community groups (e.g., religious or voluntary activities) is associated with higher cognitive scores. Improved social connections, particularly living with others, predict slower global cognitive decline, with sex-specific effects; men benefit more from structured social interactions.

### 5.3.5 Sleep Quality

The results also align with recent research that shows that sleep disturbances, including insomnia, restless sleep, and abnormal sleep duration, are significant risk factors for cognitive decline and dementia (Mander et al., 2016) (Burke et al., 2016) (Ma et al., 2020). Evidence from Burke et al. (2016) indicates that abnormal sleep (either under 6 or over 8 hours per night) and poor sleep efficiency accumulate amyloid and apolipoprotein E that accelerate neurodegeneration in the brain, increasing the risk of MCI.

### 5.3.6 Stress

On the other hand, if individuals couldn't release their stress through lifestyle intervention, communication, and social interaction, sleep/rest, or other cognitive resilience activities. Seminal multi-cohort studies demonstrate that chronic stress and persistent high levels of psychological stress, including frequent feelings of effortful daily living and chronic life strain, are associated with an increased risk of MCI and faster dementia progression. (Johansson et al., 2010) (Wilson et al., 2007).

### 5.3.7 Body Mass Index (BMI)

Our results also reports a lower late-life BMI is associated with greater risk for MCI and ADRD, while obesity may be paradoxically protective in elderly populations. Rapid declines in BMI predate, particularly values below 18.5 kg/m², clinical onset, functioning as a biomarker for heightened dementia risk (Qizilbash et al., 2015) (Guo et al., 2022) (Li et al., 2024) used a large data set ( ~2 million people age >40) longitudinal data to evidence that being underweight in middle age and old age carries an increased risk of dementia.

## 6. Conclusion

This study demonstrates that SDOH can serve as powerful predictors of cognitive performance, offering a scalable and interpretable approach for early detection of Alzheimer's disease risk. By leveraging the PREPARE Challenge dataset and implementing an SVD-based imputation pipeline alongside XGBoost modeling, we achieved superior predictive accuracy compared to existing benchmarks. The robustness of this framework in handling substantial missingness highlights its practical applicability for large, heterogeneous datasets often encountered in population-level research.

Beyond methodological advances, our analysis reveals that environmental and SDOH, such as flooring material, encapsulate SES inequities that significantly influence cognitive aging. Coupled with other key factors like SES, age, lifestyle interventions, social interaction, sleep quality, stress, and BMI, these findings underscore the multifactorial nature of cognitive health. Integrating such diverse predictors into risk models can inform targeted interventions and equitable resource allocation, ultimately contributing to more effective strategies for dementia prevention and care.